\documentclass[journal]{IEEEtran}

\usepackage{graphicx}
\usepackage{amssymb}
\usepackage{multirow}
\usepackage{algorithm}
\usepackage{algorithmic}
\usepackage{float}
\usepackage{indentfirst}
\usepackage[cmex10]{amsmath}
\usepackage{amsfonts}
\usepackage{array}
\usepackage{multirow}
\usepackage{url}
\usepackage{color}
\usepackage{mathrsfs} 
\usepackage[switch]{lineno}
\usepackage[sort&compress,numbers]{natbib}
\usepackage[pagebackref=true,colorlinks=true,urlcolor = blue,bookmarks=false,dvipdfm]{hyperref}

\begin{document}

\title{A Unified RGB-T Saliency Detection Benchmark:
Dataset, Baselines, Analysis and A Novel Approach}
\author{Chenglong Li,~~Guizhao Wang,~~Yunpeng Ma,~~Aihua Zheng,~~Bin Luo,~~and Jin Tang\thanks{The authors are with Anhui University, Hefei 230601, China. Email: jtang99029@foxmail.com.}}

\maketitle

\begin{abstract}

Despite significant progress, image saliency detection still remains a challenging task in complex scenes and environments. Integrating multiple different but complementary cues, like RGB and Thermal (RGB-T), may be an effective way for boosting saliency detection performance. The current research in this direction, however, is limited by the lack of a comprehensive benchmark. This work contributes such a RGB-T image dataset, which includes 821 spatially aligned RGB-T image pairs and their ground truth annotations for saliency detection purpose. The image pairs are with high diversity recorded under different scenes and environmental conditions, and we annotate 11 challenges on these image pairs for performing the challenge-sensitive analysis for different saliency detection algorithms. We also implement 3 kinds of baseline methods with different modality inputs to provide a comprehensive comparison platform. 

With this benchmark, we propose a novel approach, multi-task manifold ranking with cross-modality consistency, for RGB-T saliency detection. In particular, we introduce a weight for each modality to describe the reliability, and integrate them into the graph-based manifold ranking algorithm to achieve adaptive fusion of different source data. Moreover, we incorporate the cross-modality consistent constraints to integrate different modalities collaboratively. For the optimization, we design an efficient algorithm to iteratively solve several subproblems with closed-form solutions. Extensive experiments against other baseline methods on the newly created benchmark demonstrate the effectiveness of the proposed approach, and we also provide basic insights and potential future research directions for RGB-T saliency detection.
\end{abstract}


\section{Introduction}
\label{sec::introduction}
\IEEEPARstart{I}{mage} saliency detection is a fundamental and active problem in computer vision. It aims at highlighting salient foreground objects automatically from background, and has received increasing attentions due to its wide range of applications in computer vision and graphics, such as object recognition, content-aware retargeting, video compression, and image classification. Despite significant progress, image saliency detection still remains a challenging task in complex scenes and environments.

Recently, integrating RGB data and thermal data (RGB-T data) has been proved to be effective in several computer vision problems, such as moving object detection~\cite{Zhao2013Infrared,Li2016WELD} and tracking~\cite{li2016learning}. Given the potentials of RGB-T data, however, the research of RGB-T saliency detection is limited by the lack of a comprehensive image benchmark.

In this paper, we contribute a comprehensive image benchmark for RGB-T saliency detection, and the following two main aspects are considered in creating this benchmark.
\begin{itemize}

\item A good dataset should be with reasonable size, high diversity and low bias~\cite{bias11cvpr}. Therefore, we use our recording system to collect 821 RGB-T image pairs in different scenes and environmental conditions, and each image pair is aligned and annotated with ground truth. In addition, the category, size, number and spatial information of salient objects are also taken into account for enhancing the diversity and challenge, and we present some statistics of the created dataset to analyze the diversity and bias. To analyze the challenge-sensitive performance of different algorithms, we annotate 11 different challenges according to the above-mentioned factors.

\item To the best of our knowledge, RGB-T saliency detection remains not well investigated. Therefore, we implement some baseline methods to provide a comparison platform. On one hand, we regard RGB or thermal images as inputs in some popular methods to achieve single-modality saliency detection. These baselines can be utilized to identify the importance and complementarity of RGB and thermal information with comparing to RGB-T saliency detection methods. On the other hand, we concatenate the features extracted from RGB and thermal modalities together as the RGB-T feature representations, and employ some popular methods to achieve RGB-T saliency detection.

\end{itemize}

Salient object detection has been extensively studied in past decades, and numerous models and algorithms have been proposed based on different mathematical principles or priors~\cite{Harel2006Graph, Liu2011Learning, Jiang2013Salient, Yan2013Hierarchical,yu2014tc,zhu2014tc, Wang15tip, Cheng2015Global,Liang2015tc, jian2015tc, Tu-CVPR-2016}. Most of methods measured saliency by measuring local center-surround contrast and rarity of features over the entire image~\cite{Harel2006Graph, Hou2007Saliency, achanta2009frequency, Cheng2015Global}. In contrast, Gopalakrishnan et al.~\cite{Gopalakrishnan10tip} formulated the object detection problem as a binary segmentation or labelling task on a graph. The most salient seeds and several background seeds were identified by the behavior of random walks on a complete graph and a $k$-regular graph. Then, a semi-supervised learning technique was used to infer the binary labels of the unlabelled nodes. Different from it, Yang et al.~\cite{Yang2013Saliency} employed manifold ranking technique to salient object detection that requires only seeds from one class, which are initialized with either the boundary priors or foreground cues. They then extended their work with several improvements~\cite{Zhang17pami}, including multi-scale graph construction and a cascade scheme on a multi-layer representation. Based on the manifold ranking algorithms, Li et al.~\cite{Li2015Robust} generated pixel-wise saliency maps via regularized random walks ranking, and Wang et al.~\cite{Wang2016GraB} proposed a new graph model which captured local/global contrast and effectively utilized the boundary prior.

With the created benchmark, we propose a novel approach, multi-task manifold ranking with cross-modality consistency, for RGB-T saliency detection. For each modality, we employ the idea of graph-based manifold ranking~\cite{Yang2013Saliency} for the good saliency detection performance in terms of accuracy and speed. Then, we assign each modality with a weight to describe the reliability, which is capable of dealing with occasional perturbation or malfunction of individual sources, to achieve adaptive fusion of multiple modalities. To better exploit the relations among modalities, we impose the cross-modality consistent constraints on the ranking functions of different modalities to integrate them collaboratively. Considering the manifold ranking in each modality as an individual task, our method is essentially formulated as a multi-task learning problem. For the optimization, we jointly optimize the modality weights and the ranking functions of multiple modalities by iteratively solving two subproblems with closed-form solutions.

This paper makes the following three major contributions for RGB-T image saliency detection and related applications.
\begin{itemize}
  \item It creates a comprehensive benchmark for facilitating evaluating different RGB-T saliency detection algorithms. The benchmark dataset includes 821 aligned RGB-T images with the annotated ground truths, and we also present the fine-grained annotations with 11 challenges to allow us to analyze the challenge-sensitive performance of different algorithms. Moreover, we implement 3 kinds of baseline methods with different inputs (RGB, thermal and RGB-T) for evaluations. This benchmark will be available online for free academic usage~\footnote{RGB-T saliency detection benchmark's webpage:\\ \href{http://chenglongli.cn/people/lcl/journals.html} {~~~~http://chenglongli.cn/people/lcl/journals.html}.}.

  \item  It proposes a novel approach, multi-task manifold ranking with cross-modality consistency, for RGB-T saliency detection. In particular, we introduce a weight for each modality to represent the reliability, and incorporate the cross-modality consistent constraints to achieve adaptive and collaborative fusion of different source data. The modality weights and ranking function are jointly optimized by iteratively solving several subproblems with closed-form solutions.

  \item  It presents extensive experiments against other state-of-the-art image saliency methods with 3 kinds of inputs. The evaluation results demonstrate the effectiveness of the proposed approach. Through analyzing quantitative results, we further provide basic insights and identify the potentials of thermal information in RGB-T saliency detection. 

\end{itemize}

The rest of this paper is organized as follows. Sect.~\ref{sec::benchmark} introduces details of the RGB-T saliency detection benchmark. The proposed model and the associated optimization algorithm are presented in Sect.~\ref{sec::proposed_model}, and the RGB-T saliency detection approach is introduced in Sect.~\ref{sec::proposed_appraoch}. The experimental results and analysis are shown in Sect.~\ref{sec::experiments}. Sect.~\ref{sec::conclusion} concludes the paper.

\section{RGB-T Image Saliency Benchmark}
\label{sec::benchmark}
In this section, we introduce our newly created RGB-T saliency benchmark, which includes dataset with statistic analysis, baseline methods with different inputs and evaluation metrics.

\begin{figure*}[t]
\centering

\includegraphics[width =\linewidth]{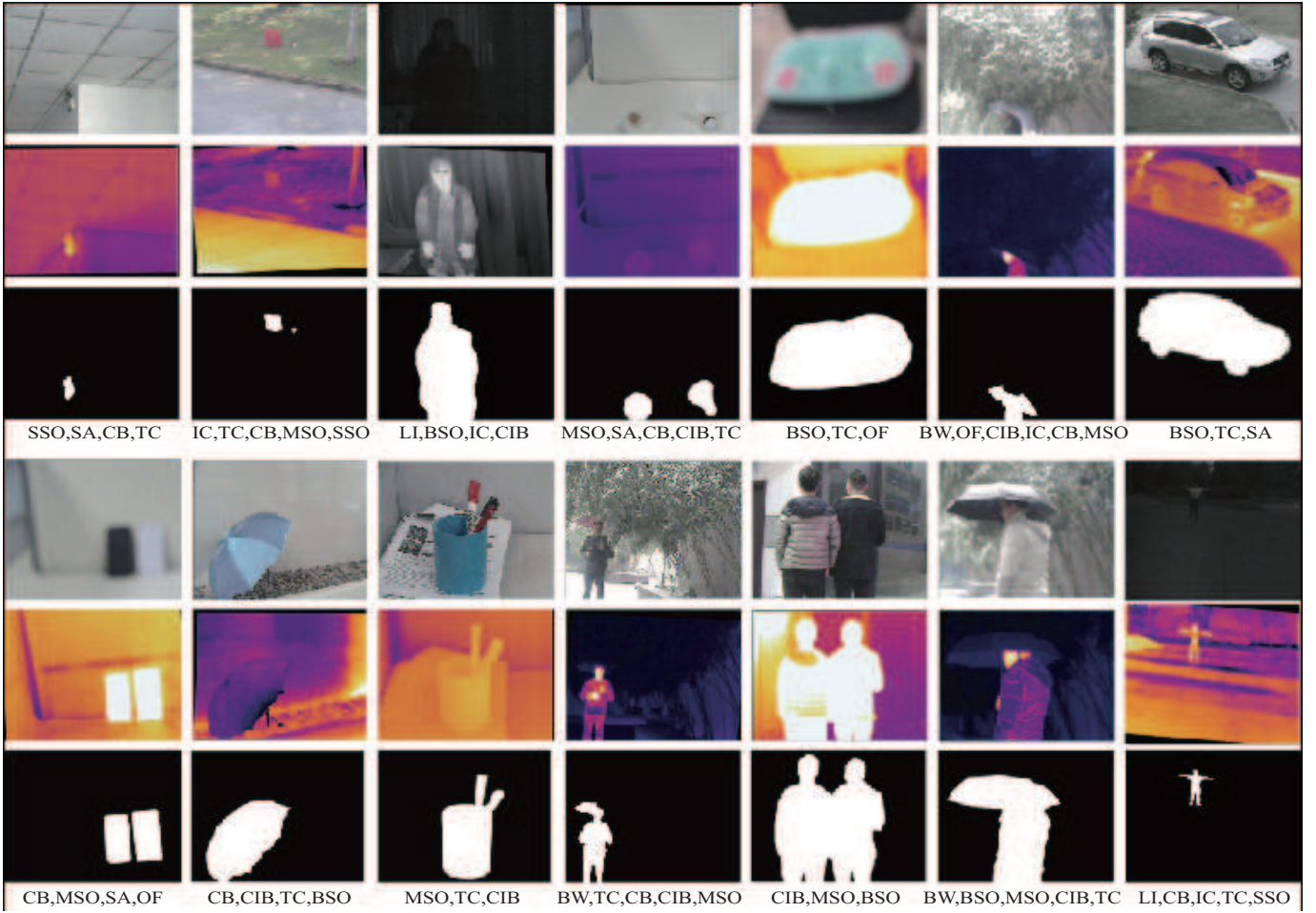}
\caption{Sample image pairs with annotated ground truths and challenges from our RGB-T dataset. }
\label{fig::sample_pairs}
\end{figure*}

\subsection{Dataset}
We collect 821 RGB-T image pairs by our recording system, which consists of an online thermal imager (FLIR A310) and a CCD camera (SONY TD-2073). For alignment, we uniformly select a number of point correspondences in each image pairs, and compute the homography matrix by the least-square algorithm. It is worth noting that this registration method can accurately align image pairs due to the following two reasons. First, we carefully choose the planar and non-planar scenes to make the homography assumption effective. Second, since two camera views are almost coincident as we made, the transformation between two views is simple. As each image pair is aligned, we annotate the pixel-level ground truth using more reliable modality. Fig.~\ref{fig::sample_pairs} shows some sample image pairs and their ground truths.

The image pairs in our dataset are recorded in approximately 60 scenes with different environmental conditions, and the category, size, number and spatial information of salient objects are also taken into account for enhancing the diversity and challenge. Specifically, the following main aspects are considered in creating the RGB-T image dataset.

\begin{figure}[t]
\centering

\includegraphics[width =\linewidth]{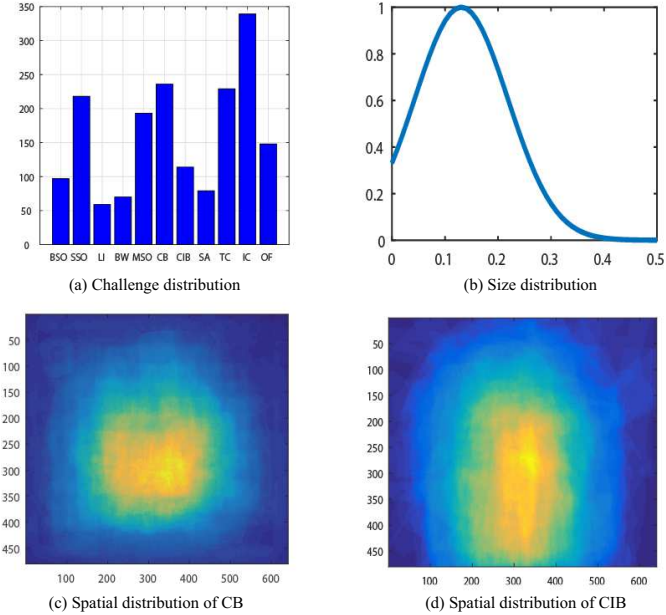}
\caption{Dataset statistics. }
\label{fig::statistics}
\end{figure}

\begin{itemize}
  \item  \emph{Illumination condition}. The image pairs are captured under different light conditions, such as sunny, snowy, and nighttime. The low illumination and illumination variation caused by different light conditions usually
bring big challenges in RGB images. 

\item  \emph{Background factor}. Two background factors are taken into account for our dataset. First, similar background to the salient objects in appearance or temperature will introduce ambiguous information. Second, it is difficult to separate objects accurately from cluttered background.

\item  \emph{Salient object attribute}. We take different attributes of salient objects, including category (more than 60 categories), size (see the size distribution in Fig.~\ref{fig::statistics} (b)) and number, into account in constructing our dataset for high diversity.

\item  \emph{Object location}. Most of methods employ the spatial information (center and boundary of an image) of the salient objects as priors, which is verified to be effective. However, some salient objects are not at center or cross image boundaries, and these situations isolate the spatial priors. We incorporate these factors into our dataset construction to bring its challenge, and Fig.~\ref{fig::statistics} presents the spatial distribution of salient objects on CB and CIB.

\end{itemize}

\begin{table}[t] \footnotesize
\centering
\caption{List of the annotated challenges of our RGB-T dataset.}\label{tb::challenges}
\begin{tabular}{|c|p{180pt}|}

\hline
\textbf{Challenge} & \textbf{Description}\\
\hline
\hline
BSO & Big Salient Object - the ratio of ground truth salient objects over image is more than 0.26.\\
SSO & Small Salient Object - the ratio of ground truth salient objects over image is less than 0.05.\\
LI & Low Illumination - the environmental illumination is low.\\
BW & Bad Weather - the image pairs are recorded in bad weathers, such as snowy, rainy, hazy and cloudy.\\
MSO & Multiple Salient Objects - the number of the salient objects in the image is more than 1.\\
CB & Center Bias - the centers of salient objects are far away from the image center. \\
CIB & Cross Image Boundary - the salient objects cross the image boundaries. \\
SA & Similar Appearance - the salient objets have similar color or shape to the background. \\
TC & Thermal Crossover - the salient objects have similar temperature to the background. \\
IC & Image Clutter - the image is cluttered. \\
OF & Out of Focus - the image is out-of-focus. \\

\hline

\end{tabular}
\end{table}

\begin{table*}[t] \footnotesize
\centering
\caption{List of the baseline methods with the used features, the main techniques and the published information. }\label{tb::baselines}
\begin{tabular}{|c||c|c|c|c|}

\hline
\textbf{Algorithm} & \textbf{Feature} & \textbf{Technique} & \textbf{Book Title} & \textbf{Year} \\
\hline
\hline

MST~\cite{Tu-CVPR-2016} & Lab \& Intensity & Minimum spanning tree & IEEE CVPR & 2016  \\
RRWR~\cite{Li2015Robust} & Lab & Regularized random walks ranking & IEEE CVPR & 2015 \\
CA~\cite{Qin2015Saliency} & Lab & Celluar Automata & IEEE CVPR & 2015  \\
GMR~\cite{Yang2013Saliency} & Lab & Graph-based manifold ranking & IEEE CVPR & 2013 \\
STM~\cite{Yan2013Hierarchical} & LUV \& Spatial information & Scale-based tree model & IEEE CVPR & 2013  \\
GR~\cite{yang2013graph} & Lab & Graph regularization & IEEE SPL & 2013  \\
NFI~\cite{Erdem2013Visual} & Lab \& Orientations \& Spatial information & Nonlinear feature integration & Journal of Vision & 2013  \\
MCI~\cite{Goferman2012Context} & Lab \& Spatial information & Multiscale context integration & IEEE TPAMI & 2012  \\
SS-KDE~\cite{Tavakoli2011Fast} & Lab & Sparse sampling and kernel density estimation & SCIA & 2011  \\
BR~\cite{rahtu2010segmenting} & Lab \& Intensity \& Motion & Bayesian reasoning & ECCV & 2010   \\
SR~\cite{seo2009static} & Lab & Self-resemblance & Journal of Vision & 2009  \\
SRM~\cite{Hou2007Saliency} & Spectrum & Spectral residual model & IEEE CVPR & 2007  \\

\hline

\end{tabular}
\end{table*}

Considering the above-mentioned factors, we annotate 11 challenges for our dataset to facilitate the challenge-sensitive performance of different algorithms. They are: big salient object (BSO), small salient object (SSO), multiple salient objects (MSO), low illumination (LI), bad weather (BW), center bias (CB), cross image boundary (CIB), similar appearance (SA), thermal crossover (TC), image clutter (IC), and out of focus (OF). Tab.~\ref{tb::challenges} shows the details, and Fig.~\ref{fig::statistics} (a) presents the challenge distribution. We will analyze the performance of different algorithms on the specific challenge using the fine-grained annotations in the experimental section.

\subsection{Baseline Methods}

To provide a comparison platform, we implement 3 kinds of baseline methods with different modality inputs. On one hand, we regard RGB or thermal images as inputs in 12 popular methods to achieve single-modality saliency detection, including MST~\cite{Tu-CVPR-2016}, RRWR~\cite{Li2015Robust}, CA~\cite{Qin2015Saliency}, GMR~\cite{Yang2013Saliency}, STM~\cite{Yan2013Hierarchical}, GR~\cite{yang2013graph}, NFI~\cite{Erdem2013Visual}, MCI~\cite{Goferman2012Context}, SS-KDE~\cite{Tavakoli2011Fast}, BR~\cite{rahtu2010segmenting}, SR~\cite{seo2009static} and SRM~\cite{Hou2007Saliency}. Tab.~\ref{tb::baselines} presents the details. These baselines can be utilized to identify the importance and complementarity of RGB and thermal information with comparing to RGB-T saliency detection methods. On the other hand, we concatenate the features extracted from RGB and thermal modalities together as the RGB-T feature representations, and employ the above-mentioned methods to achieve RGB-T saliency detection.

\subsection{Evaluation Metrics}

There exists several metrics to evaluate the agreement between subjective annotations and experimental predictions. In this work, We use (P)recision-(R)ecall curves (PR curves), $F_{0.3}$ metric and Mean Absolute Error (MAE) to evaluate all the algorithms. Given the binarized saliency map via the threshold value from 0 to 255, precision means the ratio of the correctly assigned salient pixel number in relation to all the detected salient pixel number, and recall means the ratio of the correct salient pixel number in relation to the ground truth number. Different from (P)recision-(R)ecall curves using a fixed threshold for every image, the $F_{0.3}$ metric exploits an adaptive threshold of each image to perform the evaluation. The adaptive threshold is defined as:
\begin{equation}
T= \frac{2}{w\times h}\sum\limits_{i=1}^w\sum\limits_{j=1}^h S(i,j),
\end{equation}
where $w$ and $h$ denote the width and height of an image, respectively. $S$ is the computed saliency map. The F-measure ($F$) is defined as follows with the precision ($P$) and recall ($R$) of the above adaptive threshold:
\begin{equation}
F_{\beta^2}=\frac{(1+\beta^2)\times P\times R}{\beta^2\times P + R},
\end{equation}
where we set the $\beta^2 = 0.3$ to emphasize the precision as suggested in~\cite{achanta2009frequency}. PR curves and $F_{0.3}$ metric are aimed at quantitative comparison, while MAE are better than them for taking visual comparison into consideration to estimate dissimilarity between a saliency map $S$ and the ground truth $G$, which is defined as:
\begin{equation}
MAE=\frac{1}{w \times h} \sum\limits_{i=1}^w\sum\limits_{j=1}^h |S(i,j)-G(i,j)|.
\end{equation}

\section{Graph-based Multi-Task Manifold Ranking}
\label{sec::proposed_model}
The graph-based ranking problem is described as follows: Given a graph and a node in this graph as query, the remaining nodes are ranked based on their affinities to the given query. The goal is to learn a ranking function that defines the relevance between unlabelled nodes and queries.

This section will introduce the graph-based multi-task manifold ranking model and the associated optimization algorithm. The optimized modality weights and ranking scores will be utilized for RGB-T saliency detection in next section.

\begin{figure}
  \centering
 
  \includegraphics[width=0.6\linewidth]{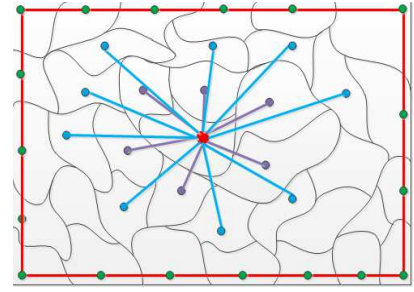}
  \caption{Illustration of graph construction.}
 \label{fig::graph_construction}
\end{figure}

\subsection{Graph Construction}
Given a pair of RGB-T images, we regard the thermal image as one of image channels, and then employ SLIC algorithm~\cite{SLIC2012} to generate $n$ non-overlapping superpixels. We take these superpixels as nodes to construct a graph $G=(V,E)$, where $V$ is a node set and $E$ is a set of undirected edges. In this work, any two nodes in $V$ are connected if one of the following conditions holds: 1) they are neighboring; 2) they share common boundaries with its neighboring node;  3) they are on the four sides of image, i.e., boundary nodes. Fig.~\ref{fig::graph_construction} shows the details. The first and second conditions are employed to capture local smoothness cues as neighboring superpixels tend to share similar appearance and saliency values. The third condition attempts to reduce the geodesic distance of similar superpixels. It is worth noting that we can explore more cues in RGB and thermal data to construct an adaptive graph that makes best use of intrinsic relationship among superpixels. We will study this issue in future work as this paper is with an emphasis on the multi-task manifold ranking algorithm.

If nodes $V_i$ and $V_j$ is connected, we assign it with an edge weight as:
\begin{equation}
\centering
\label{eq::multi-modal weight}
{\bf W}_{ij}^k=e^{-\gamma^k||{\bf c}_i^k - {\bf c}_j^k||},~k=1,2,...K,
\end{equation}
where ${\bf c}_i^k$ denotes the mean of the $i$-th superpixel in the $k$-th modality, and $\gamma$ is a scaling parameter.

\subsection{Multi-Task Manifold Ranking with Cross-Modality Consistency}

We first review the algorithm of graph-based manifold ranking that exploits the intrinsic manifold structure of data for graph labeling~\cite{Zhou04nips}. Given a superpixel feature set $X=\{{\bf x}_1,...,{\bf x}_n\}\in\mathbb{R}^{d\times n}$, some superpixels are labeled as queries and the rest need to be ranked according to their affinities to the queries, where $n$ denotes the number of superpixels. Let ${\bf s}:X\rightarrow\mathbb{R}^n$ denote a ranking function that assigns a ranking value ${\bf s}_i$ to each superpixel ${\bf x}_i$, and ${\bf s}$ can be viewed as a vector ${\bf s}=[s_1,...,s_n]^T$. In this work, we regard the query labels as initial superpixel saliency value, and ${\bf s}$ is thus an initial superpixel saliency vector. Let ${\bf y}=[{\bf y}_1,...,{\bf y}_n]^T$ denote an indication vector, in which ${\bf y}_i=1$ if ${\bf x}_i$ is a query, and ${\bf y}_i=0$ otherwise. Given $G$, the optimal ranking of queries are computed by solving the following optimization problem:
\begin{equation}
\label{eq::manifold_ranking}
\begin{aligned}
&\min_{\bf s}\frac{1}{2}(\sum_{i,j=1}^n{\bf W}_{ij}||\frac{s_i}{\sqrt{{\bf D}_{ii}}}-\frac{s_j}{\sqrt{{\bf D}_{jj}}}||^2+\mu||{\bf s}-{\bf y}||^2),
\end{aligned}
\end{equation}
where ${\bf D}=diag\{{\bf D}_{11},...,{\bf D}_{nn}\}$ is the degree matrix, and ${\bf D}_{ii}=\sum_j{\bf W}_{ij}$. $diag$ indicates the diagonal operation. $\mu$ is a parameter to balance the smoothness term and the fitting term. 

Then, we apply manifold ranking on multiple modalities, and have
\begin{equation}
\label{eq::manifold_ranking1}
\begin{aligned}
&\min_{{\bf s}^k}\frac{1}{2}(\sum_{i,j=1}^n{\bf W}^k_{ij}||\frac{s_i^k}{\sqrt{{\bf D}^k_{ii}}}-\frac{s^k_j}{\sqrt{{\bf D}^k_{jj}}}||^2+\mu||{\bf s}^k-{\bf y}||^2),\\
&k=1,2,...,K.
\end{aligned}
\end{equation}

\begin{figure}[t]
\centering

\includegraphics[width=0.45\textwidth]{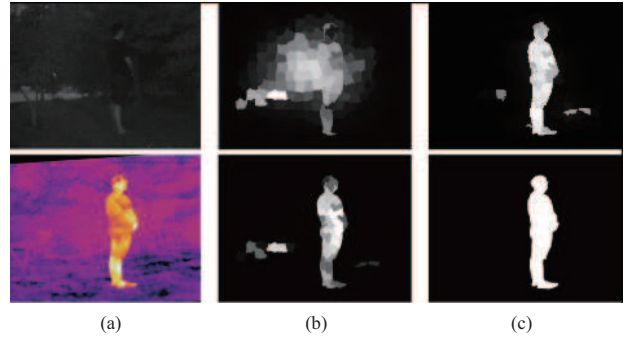}
\caption{Illustration of the effectiveness of introducing the modality weights and the cross-modality consistency. (a) Input RGB and thermal images. (b) Results of our method without modality weights and cross-modality consistency are shown in the first and second rows, respectively. (c) Our results and the corresponding ground truth. }
\label{fig::component_illustration}
\end{figure}

From Eq.~\eqref{eq::manifold_ranking1}, we can see that it inherently indicates that available modalities are independent and contribute equally. This may significantly limit the performance in dealing with occasional perturbation or malfunction of individual sources. Therefore, we propose a novel collaborative model for robustly performing salient object detection that i) adaptively integrates different modalities based on their respective modal reliabilities, ii) collaboratively computes the ranking functions of multiple modalities by incorporating the cross-modality consistent constraints. The formulation of the multi-task manifold ranking algorithm is proposed as follows:
\begin{equation}
\label{eq::multi-task_MR}
\begin{aligned}
&\min_{{\bf s}^k, {\bf r}^k}\frac{1}{2}\sum_{k=1}^{K}((r^k)^2\sum_{i,j=1}^n {\bf W}_{ij}^k||\frac{s^k_i}{\sqrt{{\bf D}_{ii}^k}}-\frac{s^k_j}{\sqrt{{\bf D}_{jj}^k}}||^2)+\\
&\mu||{\bf s}^k-{\bf y}||^2+||\Gamma \circ ({\bf 1}-{\bf r})||^2+ \lambda\sum_{k=2}^{K} ||{\bf s}^k -{\bf s}^{k-1}||^2,
\end{aligned}
\end{equation}
where $\Gamma=[\Gamma^1,...,\Gamma^K]^T$ is an adaptive parameter vector, which is initialized after the first iteration (see Alg.~\ref{alg_MMCC}), and ${\bf r}=[r^1,...,r^K]^T$ is the modality weight vector. $\circ$ denotes the element-wise product, and $\lambda$ is a balance parameter. The third term is to avoid overfitting of ${\bf r}$, and the last term is the cross-modality consistent constraints. The effectiveness of introducing these two terms is presented in Fig.~\ref{fig::component_illustration}. With some simple algebra, Eq.~\eqref{eq::multi-task_MR} can be rewritten as:
\begin{equation}
\label{eq::multi-task_MR1}
\begin{aligned}
&\min_{{\bf s}^k, {\bf r}^k}\frac{1}{2}\sum_{k=1}^{K}((r^k)^2\sum_{i,j=1}^n {\bf W}_{ij}^k||\frac{s^k_i}{\sqrt{{\bf D}_{ii}^k}}-\frac{s^k_j}{\sqrt{{\bf D}_{jj}^k}}||^2)+\\
&\mu||{\bf s}^k-{\bf y}||^2+||\Gamma \circ ({\bf 1}-{\bf r})||^2+ \lambda||{\bf CS}||_F^2,
\end{aligned}
\end{equation}
where ${\bf S}=[{\bf s}^1;{\bf s}^2;...;{\bf s}^K]\in \mathbb{R}^{nK\times 1}$, and ${\bf C}\in \mathbb{R}^{n(K-1)\times nK}$ denotes the cross-modality consistent constraint matrix, which is defined as:
\begin{equation}
\centering
\label{eq::C}
\begin{bmatrix}
{\bf I}^{2,1} &-{\bf I}^{2}& {\bf 0} & \cdots\ & {\bf 0} & {\bf 0} \\
{\bf 0} & {\bf I}^{3,2} & -{\bf I}^{3}& \cdots\ & {\bf 0} & {\bf 0} \\
 & \cdots\ & & & \cdots\ & \\
{\bf 0} & {\bf 0} & {\bf 0} & \cdots\ &{\bf I}^{K,K-1}&-{\bf I}^{K}\\
\end{bmatrix}\quad,
\end{equation}
where ${\bf I}^k$ and ${\bf I}^{k,k-1}$ are the identity matrices with the size of $n \times n$.

\begin{algorithm}[t]
\caption{Optimization Procedure to Eq.~\eqref{eq::multi-task_MR1}}\label{alg_MMCC}
\begin{algorithmic}[1]
\REQUIRE
The matrix ${\bf A}^k={\bf I}-({\bf D}^k)^{-\frac{1}{2}} {\bf W}^k ({\bf D}^k)^{-\frac{1}{2}}$, the indication vector ${\bf Y}$, and the parameters $\mu$ and $\lambda$;\\
Set $r^k=\frac{1}{K}$; $\varepsilon=10^{-4}$, $maxIter=50$.
\ENSURE
${\bf S}$, ${\bf r}$.
\FOR{$t=1:maxIter$}
\STATE Update ${\bf S}$ by Eq.~\eqref{eq::updateS};
\IF{$t==1$}
\FOR{$k=1:K$}
\STATE $\Gamma^k=\sqrt{({\bf s}^k)^T {\bf A}^k {\bf s}^k}$;
\ENDFOR
\ENDIF
\FOR{$k=1:K$}
\STATE Update $r^k$ by Eq.~\eqref{eq::updateR};
\ENDFOR
\IF{$|J_t-J_{t-1}|<\varepsilon$}
\STATE Terminate the loop.
\ENDIF
\ENDFOR
\end{algorithmic}
\end{algorithm}

\subsection{Optimization Algorithm}
We present an alternating algorithm to optimize Eq.~\eqref{eq::multi-task_MR1} efficiently, and denote
\begin{equation}
\centering
\label{eq::objective}
\begin{aligned}
&J({\bf S},{\bf r})=\frac{1}{2}\sum_{k=1}^{K}((r^k)^2\sum_{i,j=1}^n {\bf W}_{ij}^k||\frac{s^k_i}{\sqrt{{\bf D}_{ii}^k}}-\frac{s^k_j}{\sqrt{{\bf D}_{jj}^k}}||^2)+\\
&\mu||{\bf s}^k-{\bf y}||^2+||\Gamma \circ ({\bf 1}-{\bf r})||^2+ \lambda||{\bf CS}||_F^2.
\end{aligned}
\end{equation}

Given ${\bf r}$, Eq.~\eqref{eq::objective} can be written as:
\begin{equation}
\centering
\label{eq::s_subproblem}
\begin{aligned}
&J({\bf S})=\frac{1}{2}\sum_{k=1}^{K}((r^k)^2\sum_{i,j=1}^n {\bf W}_{ij}^k||\frac{s^k_i}{\sqrt{{\bf D}_{ii}^k}}-\frac{s^k_j}{\sqrt{{\bf D}_{jj}^k}}||^2)+\\
&\mu||{\bf s}^k-{\bf y}||^2+ \lambda||{\bf CS}||_F^2,
\end{aligned}
\end{equation}
and we reformulate it as follows:
\begin{equation}
\centering
\label{eq::S}
\begin{aligned}
&J({\bf S})=({\bf R\circ S})^{T} {\bf A} ({\bf R \circ S}) + \mu||{\bf S-Y}||^2 + \lambda||{\bf CS}||^2,
\end{aligned}
\end{equation}
where ${\bf Y}=[{\bf y}^1;{\bf y}^2;...;{\bf y}^K]\in \mathbb{R}^{nK\times 1}$, and ${\bf A}$ is a block-diagonal matrix defined as: ${\bf A}=diag\{{\bf A}^1,{\bf A}^2,...,{\bf A}^K\}\in\mathbb{R}^{nK\times nK}$, where ${\bf A}^k={\bf I}-({\bf D}^k)^{-\frac{1}{2}} {\bf W}^k ({\bf D}^k)^{-\frac{1}{2}}$. ${\bf R}=[r^1;...;r^1;r^2;...;r^2;...;r^K;...;r^K]\in\mathbb{R}^{nK\times 1}$. Taking the derivative of $J({\bf S})$ with respect to ${\bf S}$, we have
\begin{equation}
\centering
\label{eq::updateS}
{\bf S} = \left( \frac{{\bf R}{\bf R}^T{\bf A} + \lambda{\bf C}^T{\bf C}}{\mu} + {\bf I} \right)^{-1}{\bf Y}
\end{equation}
where ${\bf I}$ is an identity matrix with the size of $nK\times nK$.

Given ${\bf S}$, Eq.~\eqref{eq::objective} can be written as:
\begin{equation}
\centering
\label{eq::r_subproblem}
\begin{aligned}
&J(r^k)=\frac{1}{2}(r^k)^2({\bf s}^k)^T {\bf A}^k {\bf s}^k +\Gamma^k(1-r^k),
\end{aligned}
\end{equation}
and we take the derivative of $J(r^k)$ with respect to $r^k$, and obtain
\begin{equation}
\centering
\label{eq::updateR}
r^k=\frac{1}{1+\frac{({\bf s}^k)^T {\bf A}^k {\bf s}^k}{(\Gamma^k)^2} },~k=1,2,...,K.
\end{equation}

A sub-optimal optimization can be achieved by alternating between the updating of ${\bf S}$ and ${\bf r}$, and the whole algorithm is summarized in Alg.~\ref{alg_MMCC}. Although the global convergence of the proposed algorithm is not proved, we empirically validate its fast convergence in our experiments. The optimized ranking functions $\{{\bf s}^k\}$ and modality weights $\{r^k\}$ will be utilized for RGB-T saliency detection in next section.

\begin{algorithm}[t]
\caption{Proposed Approach}
\label{alg::two_stage_ranking}
\begin{algorithmic}[1]
\REQUIRE
One RGB-T image pair, the parameters $\gamma^k$, $\mu_1$, $\mu_2$ and $\lambda$. \\
\ENSURE
Saliency map $\bar{\bf s}$.
\STATE {\tt//  Graph construction}
\STATE Construct the graph with superpixels as nodes, and calculate the affinity matrix ${\bf W}^k$ and the degree matrix ${\bf D}^k$ with the parameter $\gamma^k$;
\STATE {\tt// first stage ranking}
\STATE Utilize the boundary priors to generate the background queries;

\STATE Run Alg.~\ref{alg_MMCC} with the parameters $\mu_1$ and $\lambda$ to obtain the initial saliency map ${\bf s}_{fs}$ (Eq.~\eqref{eq::initial_map});
\STATE {\tt// second stage ranking}
\STATE Compute the foreground queries using the adaptive thresholds;

\STATE Run Alg.~\ref{alg_MMCC} with the parameters $\mu_2$ and $\lambda$ to compute the final saliency map $\bar{\bf s}$ (Eq.~\eqref{eq::final_map}).
\end{algorithmic}
\end{algorithm}

\section{Two-Stage RGB-T Saliency Detection}
\label{sec::proposed_appraoch}
In this section, we present the two-stage ranking scheme for unsupervised bottom-up RGB-T saliency detection using the proposed algorithm with boundary priors and foreground queries.

\subsection{Saliency Measure}
Given an input RGB-T images represented as a graph and some salient query nodes, the saliency of each node is defined
as its ranking score computed by Eq.~\eqref{eq::multi-task_MR1}. In the conventional ranking problems, the queries are manually labelled with the ground-truth. In this work, we first employ the boundary prior widely used in other works~\cite{Yang2013Saliency,Wang15tip} to highlight the salient superpixels, and select highly confident superpixels (low ranking scores in all modalities) belonging to the foreground objects as the foreground queries. Then, we perform the proposed algorithm to obtain the final ranking results, and combine them with their modality weights to compute the final saliency map.   

\subsection{Ranking with Boundary Priors}
Based on the attention theories for visual saliency~\cite{itti1998model}, we regard the boundary nodes as background seeds (the labelled data) to rank the relevances of all other superpixel nodes in the first stage. 

Taking the bottom image boundary as an example, we utilize the nodes on this side as labelled queries and the rest as the unlabelled data, and initilize the indicator ${\bf y}$ in Eq.~\eqref{eq::multi-task_MR1}. Given ${\bf y}$, the ranking values $\{{\bf s}_b^k\}$ are computed by employing the proposed ranking algorithm, and we normalize $\{{\bf s}_b^k\}$ as $\{\hat{\bf s}_b^k\}$ with the range between 0 and 1. Similarly, given the top, left and right image boundaries, we can obtain the respective ranking values $\{\hat{\bf s}_t^k\}$, $\{\hat{\bf s}_l^k\}$, $\{\hat{\bf s}_r^k\}$. We integrate them to compute the initial saliency map for each modality in the first stage:
\begin{equation}
\centering
\label{eq::initial_map}
\begin{aligned}
&{\bf s}^k_{fs}= ({\bf 1}-\hat{\bf s}^k_b)\circ({\bf 1}-\hat{\bf s}^k_t)\circ({\bf 1}-\hat{\bf s}^k_l)\circ({\bf 1}-\hat{\bf s}^k_r),\\
&k=1,2,...,K.
\end{aligned}
\end{equation}

\begin{figure*}[t]
\centering

\includegraphics[width=\textwidth]{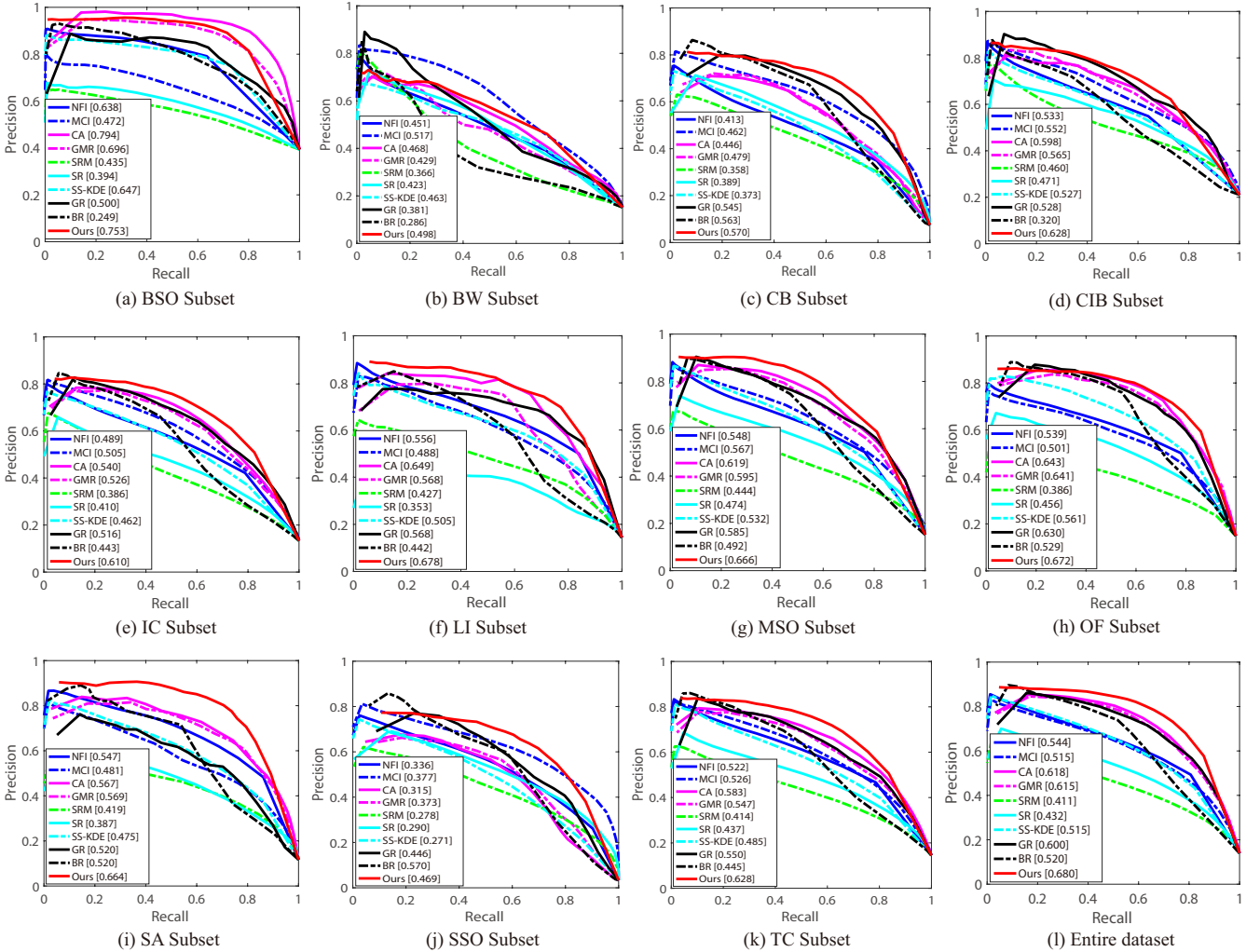}
\caption{PR curves of the proposed approach with other baseline methods with RGB-T input on the entire dataset and several subsets, where the $F_{0.3}$ values are shown in the legend. }
\label{fig::pr_curves}
\end{figure*}

\subsection{Ranking with Foreground Queries}
Given ${\bf s}^k_{fs}$ of the $k$-th modality, we set an adaptive threshold $T_k=max({\bf s}^k_{fs})-\epsilon$ to generate the foreground queries, where $max(\cdot)$ indicates the maximum operation, and $\epsilon$ is a constant, which is fixed to be 0.25 in this work. Specifically, we select the $i$-th superpixel as the foreground query of the $k$-th modality if ${\bf s}^k_{fs,i}>T_k$. Therefore, we compute the ranking values ${\bf s}^k_{ss}$ and the modality weights ${\bf r}$ in the second stage by employing our ranking algorithm. Similar to the first stage, we normalize the ranking value ${\bf s}^k_{ss}$ of the $k$-th modality as $\hat{\bf s}^k_{ss}$ with the range between 0 and 1. Finally, the final saliency map can be obtained by combining the ranking values with the modality weights:
\begin{equation}
\centering
\label{eq::final_map}
\begin{aligned}
\bar{\bf s} = \sum_{k=1}^K (r^k\hat{\bf s}^k_{ss}).
\end{aligned}
\end{equation}

\begin{table*}[t]\small
\centering
\caption{Average precision, recall, and F-measure of our method against different kinds of baseline methods on the newly created dataset. The code type and runtime (second) are also presented. The bold fonts of results indicate the best performance, and ``M'' is the abbreviation of MATLAB. }
\label{tb::PRF}
\begin{tabular}{|l||c|c|c||c|c|c||c|c|c||c|c|}
\hline
\multirow{2}{*}{\textbf{Algorithm}} & \multicolumn{3}{c||}{\textbf{RGB}} & \multicolumn{3}{c||}{\textbf{Thermal}} & \multicolumn{3}{c||}{\textbf{RGB-T}} & \multirow{2}{*}{\textbf{Code Type}} & \multirow{2}{*}{\textbf{Runtime}} \\
\cline{2-10}
 & $P$ & $R$ & $F$  & $P$ & $R$ & $F$ & $P$ & $R$ & $F$ &   	\\
\hline
  BR~\cite{rahtu2010segmenting} &\textbf{0.724}&0.260&0.411&0.648&0.413&0.488&\textbf{0.804}&0.366&0.520& M\&C++ & 8.23\\
\hline
  SR~\cite{seo2009static}     &0.425&0.523&0.377&0.361&0.587&0.362&0.484&0.584&0.432&M&1.60 \\
\hline
  SRM~\cite{Hou2007Saliency}    &0.411&0.529&0.384&0.392&0.520&0.380&0.428&0.575&0.411&M&0.76  \\
\hline
  CA~\cite{Qin2015Saliency}     &0.592&0.667&0.568&0.623&0.607&0.573&0.648&0.697&0.618&M&1.14  \\
\hline
  MCI~\cite{Goferman2012Context}    &0.526&0.604&0.485&0.445&0.585&0.435&0.547&0.652&0.515&M\&C++&21.89 \\
\hline
  NFI~\cite{Erdem2013Visual}    &0.557&0.639&0.532&0.581&0.599&0.541&0.564&0.665&0.544&M& 12.43\\
\hline
  SS-KDE~\cite{Tavakoli2011Fast}  &0.581&0.554&0.532&0.510&0.635&0.497&0.528&0.656&0.515&M\&C++& 0.94 \\
\hline
  GMR~\cite{Yang2013Saliency}    &0.644&0.603&0.587&\textbf{0.700}&0.574&\textbf{0.603}&0.694&0.624&0.615&M&1.11  \\
\hline
  GR~\cite{yang2013graph}     &0.621&0.582&0.534&0.639&0.544&0.545&0.705&0.593&0.600&M\&C++&2.43 \\
\hline
  STM~\cite{Yan2013Hierarchical}    &0.658&0.569&0.581&0.647&0.603&0.579&-&-&-&C++& 1.54\\
\hline
  MST~\cite{Tu-CVPR-2016}    &0.627&\textbf{0.739}&\textbf{0.610}&0.665&\textbf{0.655}&0.598&-&-&-&C++& 0.53 \\
\hline
  RRWR~\cite{Li2015Robust}   &0.642&0.610&0.589&0.689&0.580&0.596&-&-&-&C++& 2.99\\
\hline
  Ours   &-&-&-&-&-&-&0.716&\textbf{0.713}&\textbf{0.680}&M\&C++&1.39 \\
\hline

\end{tabular}
\end{table*}

\begin{table*}[t]\small
\centering
\caption{Average MAE Score of our method against different kinds of baseline methods on the newly created dataset. The bold fonts of results indicate the best performance. }
\label{tb::MAE}
\begin{tabular}{|l|c|c|c|c|c|c|c|c|c|c|c|c|c|c|c|}
\hline
MAE & CA & NFI & SS-KDE & GMR & GR & BR & SR & SRM & MCI & STM & MST & RRWR  & Ours \\
\hline
RGB & 0.163 & 0.126 & 0.122 &0.172 &0.197 &0.269 &0.300 &0.199 &0.211&0.194 &0.127 &0.171  &\textbf{0.109} \\
T & 0.225 &\textbf{0.124} & 0.132 &0.232 &0.199 &0.323 &0.218 &0.155 &0.176 &0.208 &0.129 &0.234 &0.141  \\
RGB-T & 0.195 & 0.125 & 0.127 &0.202 &0.199 &0.297 &0.260 &0.178 &0.195 &- &- &- &\textbf{0.107} \\
\hline

\end{tabular}
\end{table*}

\begin{figure*}[t]
\centering
\includegraphics[width=\textwidth]{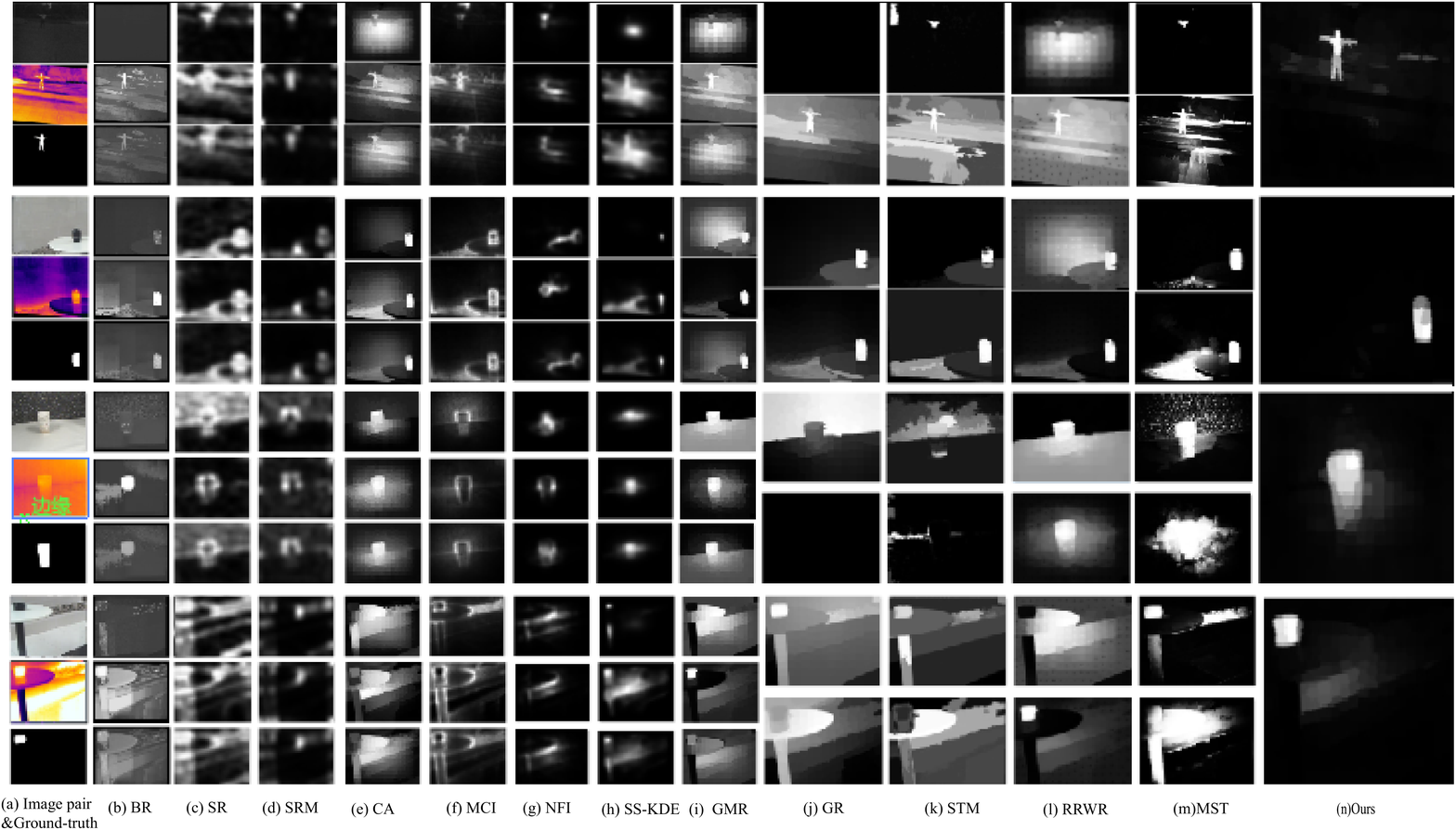}
\caption{Sample results of the proposed approach and other baseline methods with different modality inputs. (a) Input RGB and thermal image pair and their ground truth. (b-i) The results of the baseline methods with RGB, thermal and RGB-T inputs. (j-m) The results of the baseline methods with RGB and thermal inputs. (n) The results of our approach. }
\label{fig::sample_results}
\end{figure*}

\section{Experiments}
\label{sec::experiments}
In this section, we apply the proposed approach over our RGB-T benchmark and compare with other baseline methods. The source codes and result figures will be provided with the benchmark for public usage in the community.

\subsection{Experimental Settings}
For fair comparisons, we fix all parameters and other settings of our approach in the experiments, and use the default parameters released in their public codes for other baseline methods. 

In graph construction, we empirically generate $n=300$ superpixels and set the affinity parameters $\gamma^1=24$ and $\gamma^2=12$, which control the edge strength between two superpixel nodes. In Alg.~\ref{alg_MMCC} and Alg.~\ref{alg::two_stage_ranking}, we empirically set $\lambda=0.03$, $\mu_1 = 0.02$ (the first stage) and  $\mu_2 = 0.06$ (the second stage). Herein, we use bigger balance weight in the second stage than in the first stage as the refined foreground queries are more confident than the background queries according to the boundary prior.

\subsection{Comparison Results}
To justify the importance of thermal information, the complementary benefits to image saliency detection and the effectiveness of the proposed approach, we evaluate the compared methods with different modality inputs on the newly created benchmark, which has been introduced in Sect.~\ref{sec::benchmark}.

{\bf Overall performance}.
We first report the precision ($P$), recall ($R$) and F-measure ($F$) of 3 kinds of methods on entire dataset in Tab.~\ref{tb::PRF}. Herein, as the public source codes of STM, MST and RRWR are encrypted, we only run these methods on the single modality. From the evaluation results, we can observe that the proposed approach substantially outperforms all baseline methods. This comparison clearly demonstrates the effectiveness of our approach for adaptively incorporating thermal information. In addition, the thermal data are effective to boost image saliency detection and complementary to RGB data by observing the superior performance of RGB-T baselines over both RGB and thermal methods. We also report MAE of 3 kinds of methods on entire dataset in Tab.~\ref{tb::MAE}, the results of which are almost consistent with Tab.~\ref{tb::PRF}. The evaluation results further validate the effectiveness of the proposed approach, the importance of thermal information and the complementary benefits of RGB-T data. 

Fig.~\ref{fig::sample_results} shows some sample results of our approach against other baseline methods with different inputs. The evaluation results show that the proposed approach can detect the salient objects more accurate than other methods by adaptively and collaboratively leveraging RGB-T data. It is worth noting that some results using single modality are better than using RGB-T data. It is because that the redundant information introduced by the noisy or malfunction modality sometimes affects the fusion results in bad way. 

{\bf Challenge-sensitive performance}.
For evaluating RGB-T methods on subsets with different attributes (big salient object (BSO), small salient object (SSO), multiple salient objects (MSO), low illumination (LI), bad weather (BW), center bias (CB), cross image boundary (CIB), similar appearance (SA), thermal crossover (TC), image clutter (IC), and out of focus (OF), see Sect.~\ref{sec::benchmark} for details) to facilitate analysis of performance on different challenging factors, we present the PR curves in Fig.~\ref{fig::pr_curves}. From the results we can see that our approach outperforms other RGB-T methods with a clear margin on most of challenges except for BSO and BW. It validates the effectiveness of our method. In particular, for occasional perturbation or malfunction of one modality (e.g., LI, SA and TC), our method can effectively incorporate another modal information to detect salient objects robustly, justifying the complementary benefits of multiple source data. 

For BSO, CA~\cite{Qin2015Saliency} achieves a superior performance over ours, and it may attribute to CA takes the global color distinction and the global spacial distance into account for better capturing the global and local information. We can alleviate this problem by improving the graph construction that explores more relations among superpixels. For BW, most of methods have bad performance, but MCI obtains a big performance gain over ours, the second best one. It suggests that considering multiple cues, like low-level considerations, global considerations, visual organizational rules and high-level factors, can handle extremely challenges in RGB-T saliency detection, and we will integrate them into our framework to improve the robustness.

\subsection{Analysis of Our Approach}
We discuss the details of our approach by analyzing the main components, efficiency and limitations.

\begin{figure}[t]
\centering

\includegraphics[width=0.45\textwidth]{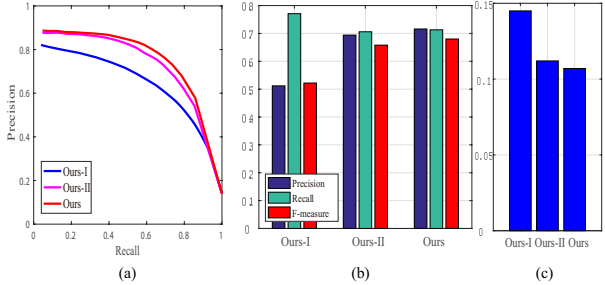}
\caption{PR curves, the representative precision, recall and F-measure and MAE of the proposed approach with its variants on the entire dataset. }
\label{fig::component}
\end{figure}

{\bf Components}.
To justify the significance of the main components of the proposed approach, we implement two special versions for
comparative analysis, including: 1) Ours-I, that removes the modality weights in the proposed ranking model, i.e., fixes $r^k=\frac{1}{K}$ in Eq.~\eqref{eq::multi-task_MR1}, and 2) Ours-II, that removes the cross-modality consistent constraints in the proposed ranking model, i.e., sets $\lambda=0$ in Eq.~\eqref{eq::multi-task_MR1}.

The PR curves, representative precision, recall and F-measure, and MAE are presented in Fig.~\ref{fig::component}, and we can draw the following conclusions. 1) Our method substantially outperforms Ours-I. This demonstrates the significance of the introduced weighted variables to achieve adaptive fusion of different source data. 2) The complete algorithm achieves superior performance than Ours-II, validating the effectiveness of the cross-modality consistent constraints.

\begin{figure}[t]
\centering

\includegraphics[width=0.3\textwidth]{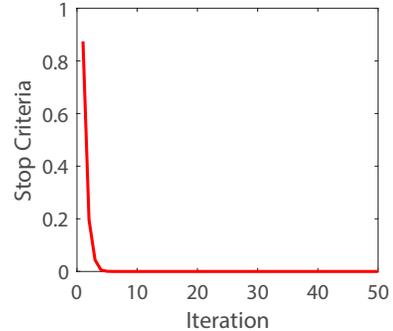}
\caption{Average convergence curve of the proposed approach on the entire dataset. }
\label{fig::convergence}
\end{figure}

{\bf Efficiency}.
Runtime of our approach against other methods is presented in Tab.~\ref{tb::PRF}. It is worth mentioning that our approach is comparable with GMR~\cite{Yang2013Saliency} mainly due to the fast optimization to the proposed ranking model.

The experiments are carried out on a desktop with an Intel i7 3.4GHz CPU and 16GB RAM, and implemented on mixing platform of C++ and MATLAB without any optimization. Fig.~\ref{fig::convergence} shows the convergence curve of the proposed approach. Although involves 5 times of the optimization to the ranking model, our method costs about 1.39 second per image pair due to the efficiency of our optimization algorithm, which converges approximately within 5 iterations. 

We also report runtime of other main procedures of the proposed approach with the typical resolution of $640\times 480$ pixels. 1) The over-segmentation by SLIC algorithm takes about 0.52 second. 2) The feature extraction takes approximately 0.24 second. 3) The first stage, including 4 ranking process, costs about 0.42 second. 4) The second stage, including 1 ranking process, spends approximately 0.14 second. The over-segmentation and the feature extraction are mostly time consuming procedure (about 55\%). Hence, through introducing the efficient over-segmentation algorithms and feature extraction implementation, we can achieve much better computation time under our approach.

\begin{figure}[t]
\centering

\includegraphics[width=0.45\textwidth]{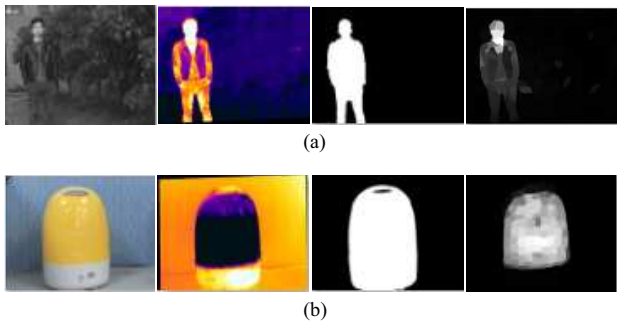}
\caption{Two failure cases of our method. The input RGB, thermal images, the ground truth and the results generated by our method are shown in (a) and (b), respectively. }
\label{fig::limitations}
\end{figure}

{\bf Limitations}.
We also present two failure cases generated by our method in Fig.~\ref{fig::limitations}. The reliable weights sometimes are wrongly estimated due to the effect of clutter background, as shown in Fig.~\ref{fig::limitations} (a), where the modality weights of RGB and thermal data are 0.97 and 0.95, respectively. In such circumstance, our method will generate bad detection results. This problem could be tackled by incorporating the measure of background clutter in optimizing reliable weights, and will be addressed in our future work. In addition to reliable weight computation, our approach has another major limitation. The first stage ranking relies on the boundary prior, and thus the salient objects are failed to be detected when they crosses image boundaries. The insufficient foreground queries obtained after the first stage usually result in bad saliency detection performance in the second stage, as shown in Fig.~\ref{fig::limitations} (b). We will handle this issue by selecting more accurate boundary queries according to other prior cues in future work.

\subsection{Discussions on RGB-T Saliency Detection}
We observe from the evaluations that integrating RGB data and thermal data will boost saliency detection performance (see Tab.~\ref{tb::PRF}). The improvements are even bigger while encountering certain challenges, i.e., low illuminations, similar appearance, image clutter and thermal crossover (see Fig.~\ref{fig::pr_curves}), demonstrating the importance of thermal information in image saliency detection and the complementary benefits from RGB-T data.

In addition, directly integrating RGB and thermal information sometimes lead to worse results than using single modality (see Fig.~\ref{fig::sample_results}), as the redundant information is introduced by the noisy or malfunction modality. We can address this issue by adaptively fusing different modal information (Our method) that can automatically determine the contribution weights of different modalities to alleviate the effects of redundant information. 

From the evaluation results, we also observe the following research potentials for RGB-T saliency detection. 1) The ensemble of multiple models or algorithms (CA~\cite{Qin2015Saliency}) can achieve robust performance. 2) Some principles are crucial for effective saliency detection, e.g., global considerations (CA~\cite{Qin2015Saliency}), boundary priors (Ours, GMR~\cite{Yang2013Saliency}, RRWR~\cite{Li2015Robust}, MST~\cite{Tu-CVPR-2016}) and multiscale context (STM~\cite{Yan2013Hierarchical}). 3) The exploitation of more relations among pixels or superpixels is important for highlighting the salient objects (STM~\cite{Yan2013Hierarchical}, MST~\cite{Tu-CVPR-2016}).

\section{Conclusion}
\label{sec::conclusion}
In this paper, we have presented a comprehensive image benchmark for RGB-T saliency detection, which includes a dataset, three kinds of baselines and four evaluation metrics. With the benchmark, we have proposed a graph-based multi-task manifold ranking algorithm to achieve adaptive and collaborative fusion of RGB and thermal data in RGB-T saliency detection. Through analyzing the quantitative and qualitative results, we have demonstrated the effectiveness of the proposed approach, and also provided some basic insights and potential research directions for RGB-T saliency detection. Our future works will focus on the following aspects: 1) We will expand the benchmark to a larger one, including a larger dataset with more challenging factors and more popular baseline methods. 2) We will improve the robustness of our approach by studying other prior models~\cite{lin2016tc} and graph construction.

\bibliographystyle{IEEEtran}
\bibliography{bibfile}

\end{document}